\documentclass[final]{cvpr}

\usepackage{times}
\usepackage{epsfig}
\usepackage{graphicx}
\usepackage{amsmath}
\usepackage{amssymb}
\usepackage{multirow}
\usepackage{url}
\usepackage{algorithm}
\usepackage[noend]{algpseudocode}
\usepackage{subcaption}
\usepackage{graphicx}

\usepackage{lipsum}
\newcommand\blfootnote[1]{%
  \begingroup
  \renewcommand\thefootnote{}\footnote{#1}%
  \addtocounter{footnote}{-1}%
  \endgroup
}
\usepackage{booktabs}
\usepackage[table]{xcolor}
\definecolor{lightgray}{gray}{0.9}
\definecolor{lightblue}{rgb}{0.93,0.95,1.0}
\definecolor{darkgreen}{rgb}{0.0,0.6,0.0}
\definecolor{blue}{rgb}{1, 0, 0}

\newcommand{\minisection}[1]{\vspace{1mm}\noindent{\textbf{#1}.}}
\newenvironment{tight_itemize}{
\begin{itemize}
  \setlength{\topsep}{0pt}
  \setlength{\itemsep}{2pt}
  \setlength{\parskip}{0pt}
  \setlength{\parsep}{0pt}
}{\end{itemize}}
\usepackage{pifont}
\newcommand{\xmark}{\text{\ding{55}}}
\newcommand{\cmark}{\ding{51}}
\newcommand{\norm}[1]{\left\lVert#1\right\rVert}
\usepackage[pagebackref=true,breaklinks=true,colorlinks,bookmarks=false]{hyperref}

\begin{document}

\title{img2pose: Face Alignment and Detection via 6DoF, Face Pose Estimation\vspace{-3mm}}

\author{Vítor Albiero$^{1,*}$, 
Xingyu Chen$^{2,*}$, Xi Yin$^2$, Guan Pang$^2$, Tal Hassner$^2$\\
$^1$University of Notre Dame\\
$^2$Facebook AI\\
}

\twocolumn[{%
\maketitle
\renewcommand\twocolumn[1][]{#1}%
\begin{center}
    \includegraphics[width=.96\textwidth,height=50mm,clip,trim=0mm 0mm 0mm 0mm]{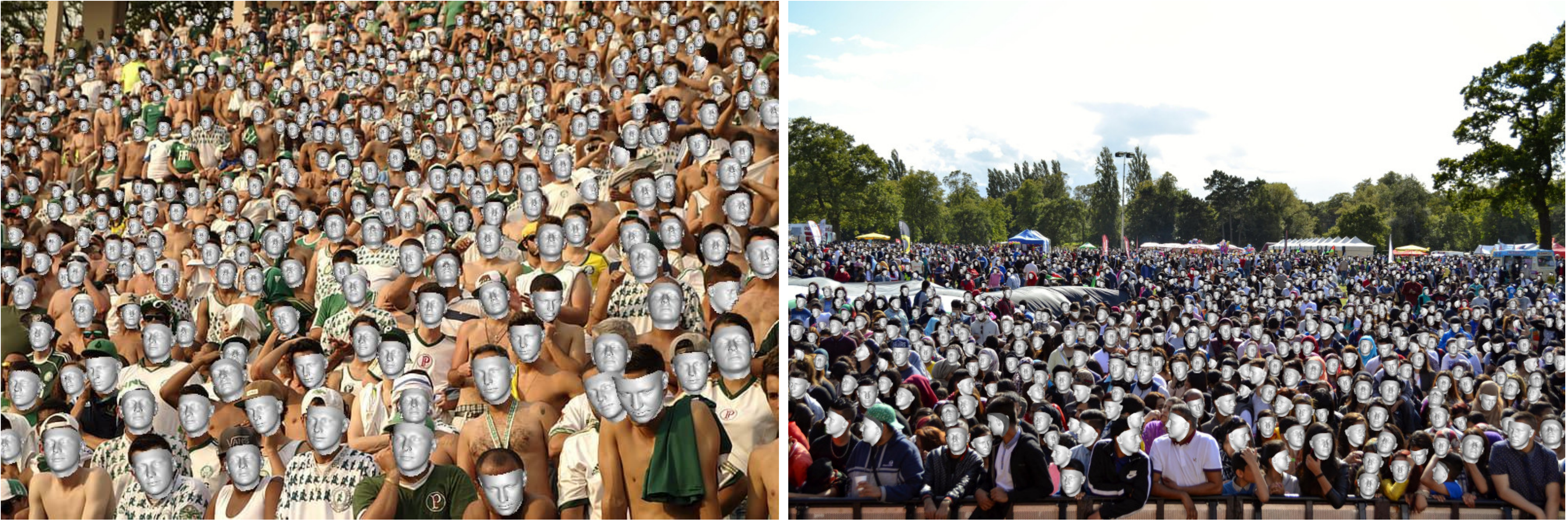}
    \captionof{figure}{We estimate the 6DoF rigid transformation of a 3D face (rendered in silver), aligning it with even the tiniest faces, {\em without face detection or facial landmark localization}. Our estimated 3D face locations are rendered by descending distances from the camera, for coherent visualization. For more qualitative results, see appendix.}
    \label{fig:teaser}%}
\end{center}%
}]

\begin{abstract}
\blfootnote{$^*$ Joint first authorship.}
\blfootnote{All experiments reported in this paper were performed at the University of Notre Dame.}We propose real-time, six degrees of freedom (6DoF), 3D face pose estimation without face detection or landmark localization. We observe that estimating the 6DoF rigid transformation of a face is a simpler problem than facial landmark detection, often used for 3D face alignment. In addition, 6DoF offers more information than face bounding box labels. We leverage these observations to make multiple contributions: (a) We describe an easily trained, efficient, Faster R-CNN--based model which regresses 6DoF pose for all faces in the photo, without preliminary face detection. (b) We explain how pose is converted and kept consistent between the input photo and arbitrary crops created while training and evaluating our model. (c) Finally, we show how face poses can replace detection bounding box training labels. Tests on AFLW2000-3D and BIWI show that our method runs at real-time and outperforms state of the art (SotA) face pose estimators. Remarkably, our method also surpasses SotA models of comparable complexity on the WIDER FACE detection benchmark, despite not been optimized on bounding box labels.\vspace{-5mm}
\end{abstract}

\section{Introduction}
{\em Face detection} is the problem of positioning a box to bound each face in a photo. {\em Facial landmark detection} seeks to localize specific facial features: e.g., eye centers, tip of the nose. Together, these two steps are the cornerstones of many face-based reasoning tasks, most notably recognition~\cite{deng2019arcface,masi2017rapid,masi2016we,masi2019face,wang2018cosface,wolf2011face} and 3D reconstruction~\cite{feng2018joint,hernandez2017accurate, tuan2017regressing,tuan2018extreme}. Processing typically begins with face detection followed by landmark detection in each detected face box. Detected landmarks are matched with corresponding ideal locations on a reference 2D image or a 3D model, and then an alignment transformation is resolved using standard means~\cite{posit, epnp}. The terms {\em face alignment} and landmark detection are thus sometimes used interchangeably~\cite{browatzki20203fabrec,dapogny2019decafa,kumar2020luvli}.

Although this approach was historically successful, it has drawbacks. Landmark detectors are often optimized to the particular nature of the bounding boxes produced by specific face detectors. Updating the face detector therefore requires re-optimizing the landmark detector~\cite{bulat2017far,feng2017face,merget2018robust,yan2013learn}. More generally, having two successive components implies separately optimizing two steps of the pipeline for accuracy and -- crucially for faces -- fairness~\cite{albiero2020skin, albiero2020training, issues_skin_2020}. In addition, SotA detection and pose estimation models can be computationally expensive (e.g., ResNet-$152$ used by the full RetinaFace~\cite{retinaface} detector). This computation accumulates when these steps are applied serially. Finally, localizing the standard 68 face landmarks can be difficult for tiny faces such as those in Fig.~\ref{fig:teaser}, making it hard to estimate their poses and align them. To address these concerns, we make the following key observations:

\minisection{Observation 1: 6DoF pose is easier to estimate than detecting landmarks} Estimating 6DoF pose is a 6D regression problem, obviously smaller than even 5-point landmark detection (5$\times$2D landmarks $=$ 10D), let alone standard 68 landmark detection ($=$136D). Importantly, pose captures the {\em rigid transformation} of the face. By comparison, landmarks entangle this rigid transformation with non-rigid facial deformations and subject-specific face shapes. 

This observation inspired many to recently propose skipping landmark detection in favor of direct pose estimation~\cite{chang2018expnet,chang2019deep,faceposenet,kuhnke2019deep,mustikovela2020self,hopenet,fsanet}. These methods, however, estimate poses for detected faces. By comparison, we aim to estimate poses without assuming that faces were already detected.

\minisection{Observation 2: 6DoF pose labels capture more than just bounding box locations} Unlike angular, 3DoF pose estimated by some~\cite{quatnet, hpe, hopenet, fsanet}, 6DoF pose can be converted to a 3D-to-2D projection matrix. Assuming a known intrinsic camera parameters, pose can therefore align a 3D face with its location in the photo~\cite{hassner2015effective}. Hence, pose already captures the location of the face in the photo. Yet, for the price of two additional scalars (6D pose vs. four values per box), 6DoF pose also provides information on the 3D position and orientation of the face. This observation was recently used by some, most notably, RetinaFace~\cite{retinaface}, to improve detection accuracy by proposing multi-task learning of bounding box and facial landmarks. We, instead, combine the two in the single goal of directly regressing 6DoF face pose. 

\vspace{3mm}

We offer a novel, easy to train, real-time solution to 6DoF, 3D face pose estimation, without requiring face detection (Fig.~\ref{fig:teaser}). We further show that predicted 3D face poses can be converted to obtain accurate 2D face bounding boxes with only negligible overhead, thereby providing face detection as a byproduct. Our method regresses 6DoF pose in a Faster R-CNN--based framework~\cite{faster_rcnn}. We explain how poses are estimated for {\em ad-hoc} proposals. To this end, we offer an efficient means of converting poses across different image crops (proposals) and the input photo, keeping ground truth and estimated poses consistent. In summary, we offer the following contributions. 
\begin{tight_itemize}
\item We propose a novel approach which estimates 6DoF, 3D face pose for all faces in an image directly, and without a preceding face detection step.
\item We introduce an efficient pose conversion method to maintain consistency of estimates and ground-truth poses, between an image and its ad-hoc proposals. 
\item We show how generated 3D pose estimates can be converted to accurate 2D bounding boxes as a byproduct with minimal computational overhead. 
\end{tight_itemize}
Importantly, all the contributions above are agnostic to the underlying Faster R-CNN--based architecture. The same techniques can be applied with other detection architectures to directly extract 6DoF, 3D face pose estimation, without requiring face detection.

Our model uses a small, fast, ResNet-$18$~\cite{resnet} backbone and is trained on the WIDER FACE~\cite{wider_face} training set with a mixture of weakly supervised and human annotated ground-truth pose labels.  We report SotA accuracy with real-time inference on both AFLW2000-3D~\cite{3ddfa} and BIWI~\cite{biwi}. We further report face detection accuracy on WIDER FACE~\cite{wider_face}, which outperforms models of comparable complexity by a wide margin. Our implementation and data are publicly available from:~\url{http://github.com/vitoralbiero/img2pose}.

\section{Related work}
\noindent{\bf Face detection} Early face detectors used hand-crafted features~\cite{HOG, edgeori, haar}. Nowadays, deep learning is used for its improved accuracy in detecting general objects~\cite{faster_rcnn} and faces~\cite{retinaface, asfd}. Depending on whether region proposal networks are used, these methods can be classified into {\em single-stage} methods~\cite{ssd, redmon2016yolo9000, yolov3} and {\em two-stage} methods~\cite{faster_rcnn}.

Most single-stage methods~\cite{dsfd, ssh, pyramidbox, s3fd} were based on the Single Shot MultiBox Detector (SSD)~\cite{ssd}, and focused on detecting small faces. For example, S$^3$FD~\cite{s3fd} proposed a scale-equitable framework with a scale compensation anchor matching strategy. PyramidBox~\cite{pyramidbox} introduced an anchor-based context association method that utilized contextual information.

Two-stage methods~\cite{facerfcn, fdnet} are typically based on Faster R-CNN~\cite{faster_rcnn} and R-FCN~\cite{rfcn}. FDNet~\cite{fdnet}, for example, proposed multi-scale and voting ensemble techniques to improve face detection. Face R-FCN~\cite{facerfcn} utilized a novel position-sensitive average pooling on top of R-FCN.

\minisection{Face alignment and pose estimation}
Face pose is typically obtained by detecting facial landmarks and then solving Perspective-n-Point (PnP) algorithms~\cite{posit, epnp}. Many landmark detectors were proposed, both conventional~\cite{burgos2013robust, cao2014face, cootes1998active, liu2007generic} and deep learning--based~\cite{bulat2017far, sun2013deep, wu2017facial, zhu2012face} and we refer to a recent survey~\cite{landmark_survey} on this topic for more information. Landmark detection methods are known to be brittle~\cite{chang2019deep, faceposenet}, typically requiring a prior face detection step and relatively large faces to position all landmarks accurately.

\begin{figure}[!t]
    \centering{
        \includegraphics[width=.48\textwidth]{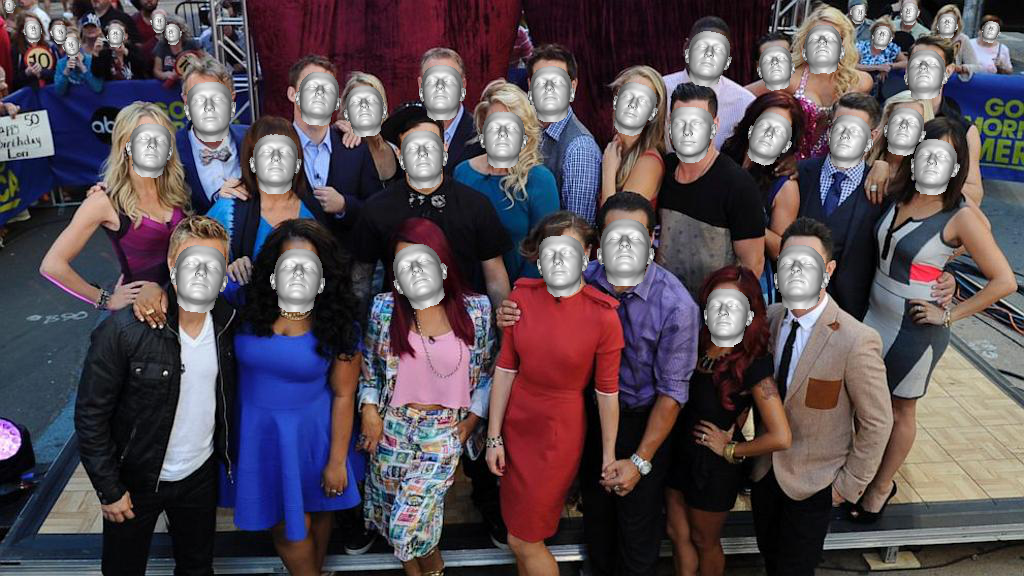}
        \includegraphics[clip,trim=0cm 0cm 0mm 0cm,width=.48\textwidth]{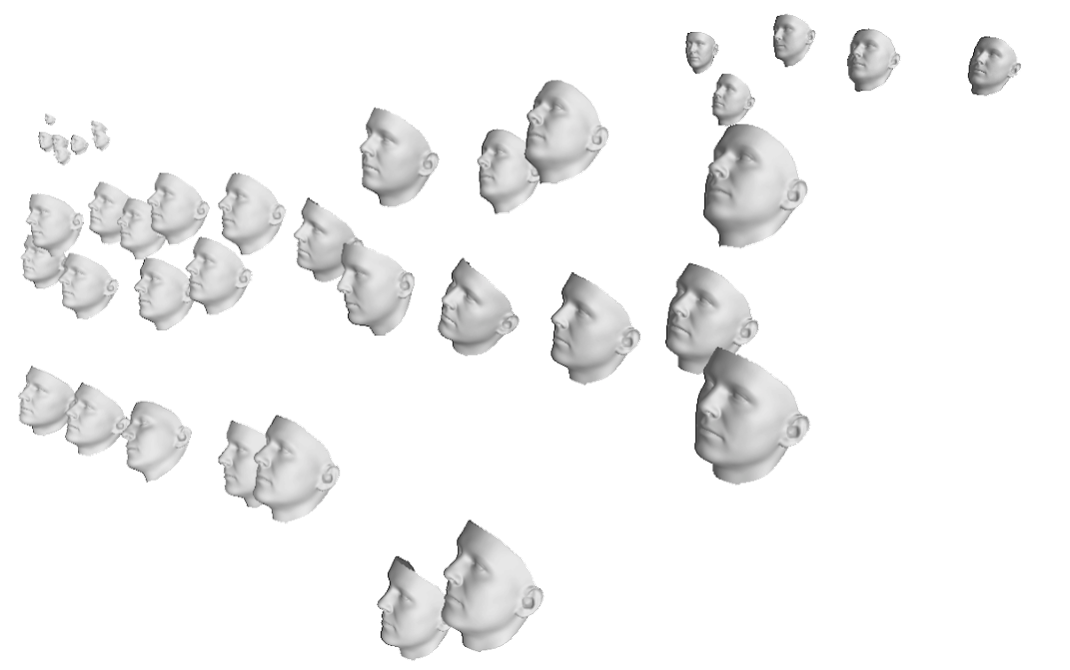}
    \caption{The 6DoF face poses estimated by our img2pose capture the positions of faces in the photo (top) and their 3D scene locations (bottom). See also Fig.~\ref{fig:analysis} for a visualization of the 3D positions of all faces in WIDER FACE (val.).\vspace{-3mm}}
    \label{fig:render3D}
    }
\end{figure}

A growing number of recent methods recognize that deep learning offers a way of directly regressing the face pose, in a {\em landmark-free} approach. Some directly estimated the 6DoF face pose from a face bounding box~\cite{chang2018expnet,chang2019deep,faceposenet,kuhnke2019deep,mustikovela2020self,hopenet,fsanet}. The impact of these {\em landmark free} alignment methods on downstream face recognition accuracy was evaluated and shown to improve results compared with landmark detection methods~\cite{chang2019deep, faceposenet}. HopeNet~\cite{hopenet} extended these methods by training a network with multiple losses, showing significant performance improvement. FSA-Net~\cite{fsanet} introduced a feature aggregation method to improve pose estimation. 
Finally, QuatNet~\cite{quatnet} proposed a Quaternion-based face pose regression framework which claims to be more effective than Euler angle-based methods. 
All these methods rely on a face detection step, prior to pose estimation whereas our approach collapses these two to a single step.

Some of the methods listed above only regress 3DoF angular pose: the face yaw, pitch, and roll~\cite{hopenet, fsanet} or rotational information~\cite{quatnet}. For some use cases, this information suffices. Many other applications, however, including face alignment for recognition~\cite{hassner2015effective,masi2017rapid,masi2016we,masi2019face,wang2018cosface,wolf2011face}, 3D reconstruction~\cite{feng2018joint, tuan2017regressing,tuan2018extreme}, face manipulation~\cite{nirkin2019fsgan,nirkin2018face,nirkin2020deepfake}, also require the translational components of a full 6DoF pose. Our img2pose model, by comparison, provides full 6DoF face pose for every face in the photo (Fig.~\ref{fig:render3D}). 

Finally, some noted that face alignment is often performed along with other tasks, such as face detection, landmark detection, and 3D reconstruction. They consequently proposed solving these problems together in a {\em multi-task manner}. Some early examples of this approach predate the recent rise of deep learning~\cite{osadchy2007synergistic,osadchy2004synergistic}. More recent methods add face pose estimation or landmark detection heads to a face detection network~\cite{chang2019deep, kepler, hyperface, ranjan2017all, zhu2012face}. It is unclear, however, if adding these tasks together improves or hurts the accuracy of the individual tasks. Indeed, evidence suggesting the latter is growing~\cite{lu2017fully,tran2019transferability,zhao2018modulation}. We leverage the observation that pose estimation already encapsulates face detection, thereby requiring only 6DoF pose as a single supervisory signal.

\section{Proposed method}\label{sec:method}
Given an image ${\bf I}$, we estimate 6DoF pose for each face, $i$ appearing in ${\bf I}$. We use ${{\bf{h}}_i}\in\mathbb{R}^6$ to denote each face pose: 
\begin{equation}
{{\bf{h}}_i} = (r_x, r_y, r_z, t_x, t_y, t_z),\label{eq:pose}
\end{equation}
where $(r_x, r_y, r_z)$ represent a rotation vector \cite{trucco1998introductory} and $(t_x, t_y, t_z)$ is the 3D face translation.

It is well known that a 6DoF face pose, $\bf{h}$, can be converted to an extrinsic camera matrix for projecting a 3D face to the 2D image plane~\cite{forsyth2002computer,szeliski2010computer}. Assuming known intrinsic camera parameters, the 3D face can then be aligned with a face in the photo~\cite{hassner2013viewing,hassner2015effective}. To our knowledge, however, previous work never leveraged this observation to propose replacing training for face bounding box detection with 6DoF pose estimation.

Specifically, assume a 3D face shape represented as a triangulated mesh. Points on the 3D face surface can be projected down to the photo using the standard pinhole model~\cite{hartley2003multiple}:
\begin{equation}
[\mathbf{Q}, \mathbf{1}]^T 	\sim \mathbf{K}[\mathbf{R}, \mathbf{t}][\mathbf{P}, \mathbf{1}]^T,
\label{eqn:proj}
\end{equation}
 where $\mathbf{K}$ is the intrinsic matrix (Sec.~\ref{sec:pose_conversion}), $\mathbf{R}$ and $\mathbf{t}$ are the 3D rotation matrix and translation vector, respectively, obtained from $\bf{h}$ by standard means~\cite{forsyth2002computer,szeliski2010computer}, and $\mathbf{P} \in \mathbb{R}^{3 \times n}$ is a matrix representing $n$ 3D points on the surface of the 3D face shape. Finally, $\mathbf{Q} \in \mathbb{R}^{2 \times n}$ is the matrix representation of 2D points projected from 3D onto the image. 

We use Eq.~\eqref{eqn:proj} to generate our qualitative figures, aligning the 3D face shape with each face in the photo (e.g., Fig.~\ref{fig:teaser}). Importantly, given the projected 2D points, $\mathbf{Q}$, a face detection bounding box can simply be obtained by taking the bounding box containing these 2D pixel coordinates. 

\begin{figure}[!t]
    \centering{
        \includegraphics[width=1\columnwidth]{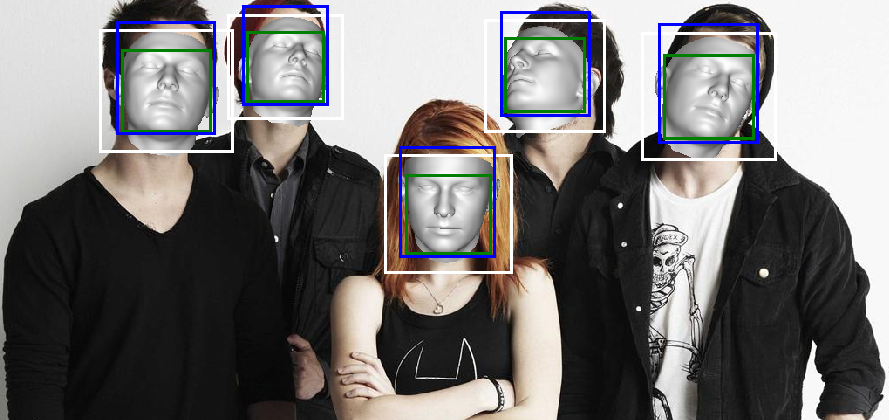}
    \caption{Bounding boxes generated using predicted poses. White bounding boxes generated with a loose setting, green with very tight setting, and blue with a less tight setting and forehead expansion (which is located through the pose).\vspace{-5mm}}
    \label{fig:bb_generation}
    }
\end{figure}

It is worth noting that this approach provides better control over bounding box looseness and shapes, as shown in Fig.~\ref{fig:bb_generation}. Specifically, because the pose aligns a 3D shape with known geometry to a face region in the image, we can choose to modify face bounding boxes sizes and shapes to match our needs, e.g., including more of the forehead by expanding the box in the correct direction, invariant of pose.

\subsection{Our img2pose network}
\label{sec:network}
\begin{figure*}[ht]
    \centering{
    \includegraphics[width=1\textwidth]{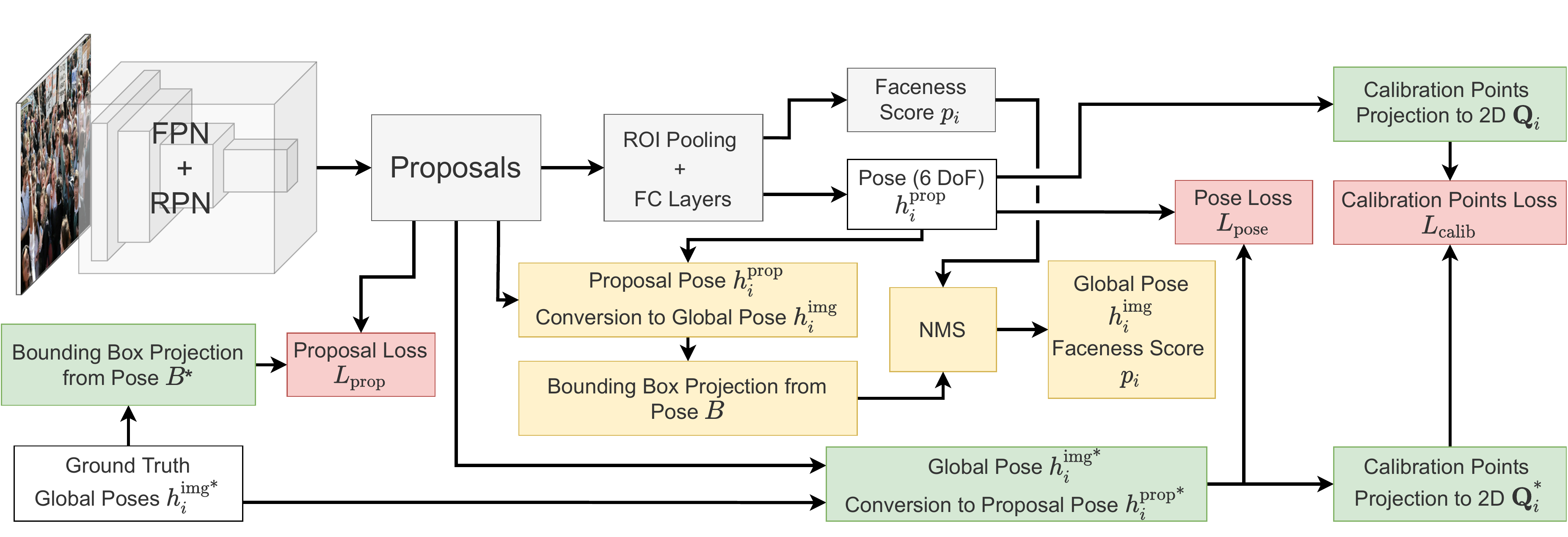}\vspace{-2mm}
    \caption{Overview of our proposed method. Components that only appear in training time are colored in green and red, and components that only appear in testing time are colored in yellow. Gray color denotes default components from Faster R-CNN with FPN~\cite{fpn, faster_rcnn}. Please see Sec.~\ref{sec:method} for more details.\vspace{-5mm}}
    \label{fig:network}
    }
\end{figure*}

We regress 6DoF face pose directly, based on the observation above that face bounding box information is already {\em folded} into the 6DoF face pose. Our network structure is illustrated in Fig.~\ref{fig:network}. Our network follows a two-stage approach based on Faster R-CNN~\cite{faster_rcnn}. The first stage is a region proposal network (RPN) with a feature pyramid~\cite{fpn}, which proposes potential face locations in the image. 

Unlike the standard RPN loss, $L_{rpn}$, which uses ground-truth bounding box labels, we use projected bounding boxes, $\mathbf{B^*}$, obtained from the 6DoF ground-truth pose labels using Eq.~\eqref{eqn:proj} (see Fig.~\ref{fig:network}, $L_{prop}$). As explained above, by doing so, we gain better consistency in the facial regions covered by our bounding boxes, $\mathbf{B^*}$. Other aspects of this stage are similar to those of the standard Faster R-CNN~\cite{faster_rcnn}, and we refer to their paper for technical details.

The second stage of our img2pose extracts features from each proposal with region of interest (ROI) pooling, and then passes them to two different heads: a standard face/non-face ({\em faceness}) classifier and a novel 6DoF face pose regressor (Sec.~\ref{sec:posereg}).

\subsection{Pose label conversion}
\label{sec:pose_conversion}
Two stage detectors rely on proposals -- ad hoc image crops -- as they train and while being evaluated. The pose regression head is provided with features extracted from proposals, not the entire image, and so does not have information required to determine where the face is located in the entire photo. This information is necessary because the 6DoF pose values are directly affected by image crop coordinates. For instance, a crop tightly matching the face would imply that the face is very close to the camera (small $t_z$ in Eq.~\eqref{eq:pose}) but if the face appears much smaller in the original photo, this value would change to reflect the face being much farther away from the camera.   

We therefore propose adjusting poses for different image crops, maintaining consistency between proposals and the entire photo. Specifically, for a given image crop we define a {\em crop camera intrinsic matrix}, $\mathbf{K}$, simply as:
\begin{equation}
    \mathbf{K} = \begin{bmatrix}
    f & 0 & c_x\\
    0 & f & c_y \\
    0 & 0 & 1
    \end{bmatrix}\label{eq:intrinsic}
\end{equation}
Here, $f$ equals the face crop height plus width, and $c_x$ and $c_y$ are the $x,y$ coordinates of the crop center. Pose values are then converted between local (crop) and global (entire photo) coordinate frames, as follows. 

Let matrix $\mathbf{K}_{img}$ be the projection matrix for the entire image, where $w$ and $h$ are the image width and height respectively, and $\mathbf{K}_{box}$ be the projection matrix for an arbitrary face crop (e.g., proposal), defined by $\mathbf{B} = (x, y, w_{bb}, h_{bb})$, where $w_{bb}$ and $h_{bb}$ are the face crop width and height respectively, and $c_x$ and $c_y$ are the $x,y$ coordinates of the face crop's center. We define these matrices as:
\begin{equation}
    \mathbf{K}_{box} = \begin{bmatrix}
    w+h & 0 & c_x + x\\
    0 & w+h & c_y + y\\
    0 & 0 & 1
    \end{bmatrix}\label{eq:cropproj}
\end{equation}

\begin{equation}
    \mathbf{K}_{img} = \begin{bmatrix}
    w+h & 0 & w/2\\
    0 & w+h & h/2 \\
    0 & 0 & 1
    \end{bmatrix}\label{eq:imgproj}
\end{equation}

\minisection{Converting pose from local to global frames} Given a pose, $\mathbf{h}^{prop}$, in a face crop coordinate frame, $\mathbf{B}$, intrinsic matrix, $\mathbf{K}_{img}$, for the entire image, intrinsic matrix, $\mathbf{K}_{box}$, for a face crop, we apply the method described in Algorithm~\ref{alg:poseconversion} to convert $\mathbf{h}^{prop}$ to $\mathbf{h}^{img}$ (see Fig.~\ref{fig:network}).

\begin{algorithm}[h]
\caption{Local to global pose conversion}
\label{alg:poseconversion}
\begin{algorithmic}[1]
	\Procedure{pose\_convert}{$\mathbf{h}^{prop}$, $\mathbf{B}$, $\mathbf{K}_{box}$, $\mathbf{K}_{img}$}
	    \State $f \leftarrow w + h$
	    \State $t_z = t_z * f / (w_{bb} + h_{bb})$
	    \State $\mathbf{V} = \mathbf{K}_{box} [t_x, t_y, t_z]^T$
	    \State $[t'_x, t'_y, t'_z]^T = (\mathbf{K}_{img})^{-1} \mathbf{V}$
	    \State $\mathbf{R} = \text{rot\_vec\_to\_rot\_mat}([r_x, r_y, r_z])$
	    \State $\mathbf{R'} = (\mathbf{K}_{img})^{-1} \mathbf{K}_{box} \mathbf{R} $
	    \State $(r'_x, r'_y, r'_z) = \text{rot\_mat\_to\_rot\_vec}(\mathbf{R'})$
	    \State \textbf{return} $\mathbf{h}^{img} = (r'_x, r'_y, r'_z, t'_x, t'_y, t'_z)$  
	\EndProcedure
\end{algorithmic}
\end{algorithm}

Briefly, Algorithm~\ref{alg:poseconversion} has two steps. First, in lines 2--3, we rescale the pose. Intuitively this step adjusts the camera to view the entire image, not just a crop. Then, in steps 4--8, we translate the focal point, adjusting the pose based on the difference of focal point locations, between the crop and the image. Finally, we return a 6DoF pose relative to the image intrinsic, $\mathbf{K}_{img}$. 
The functions \text{rot\_vec\_to\_rot\_mat($\cdot$)} and \text{rot\_mat\_to\_rot\_vec($\cdot$)} are standard conversion functions between rotation matrices and rotation vectors~\cite{hartley2003multiple, trucco1998introductory}.
Please see Appendix~\ref{sec:append:poseconvert} for more details on this conversion.

\minisection{Converting pose from global to local frames}
To convert pose labels, $\mathbf{h}^{img}$, given in the image coordinate frame, to local crop frames, $\mathbf{h}^{prop}$, we apply a process similar to Algorithm~\ref{alg:poseconversion}. Here, $\mathbf{K}_{img}$ and $\mathbf{K}_{box}$ change roles, and scaling is applied last. 
We provide details of this process in Appendix~\ref{sec:append:poseconvert} (see Fig.~\ref{fig:network}, $\mathbf{h}_i^{img*}$).
This conversion is an important step, since, as previously mentioned, proposal crop coordinates vary constantly as the method is trained and so ground-truth pose labels given in the image coordinate frame must be converted to match these changes.

\subsection{Training losses}
\label{sec:posereg}
We simultaneously train both the face/non-face classifier head and the face pose regressor. For each proposal, the model employs the following multi-task loss $L$. 
\begin{equation}
\begin{split}
L & = L_{cls}(p_i, p^*_i) + p^*_i \cdot L_{pose}(\mathbf{h}_i^{prop}, \mathbf{h}_i^{prop*}) \\
  & + p^*_i \cdot L_{calib}(\mathbf{Q}^c_i, \mathbf{Q}^{c*}_i),
\end{split}
\end{equation}
which includes these three components: 

\minisection{(1) Face classification loss} We use standard binary cross-entropy loss, $L_{cls}$, to classify each proposal, where $p_i$ is the probability of proposal $i$ containing a face and $p^*_i$ is the ground-truth binary label ($1$ for face and $0$ for background). These labels are determined by calculating the intersection over union (IoU) between each proposal and the ground-truth projected bounding boxes. For negative proposals which do not contain faces, ($p^*_i = 0$), $L_{cls}$ is the only loss that we apply. For positive proposals, ($p^*_i=1$), we also evaluate the two novel loss functions described below.

\minisection{(2) Face pose loss} This loss directly compares a 6DoF face pose estimate with its ground truth. Specifically, we define
\begin{equation}
    L_{pose}(\mathbf{h}_i^{prop}, \mathbf{h}_i^{prop*}) = \norm{\mathbf{h}_i^{prop} - \mathbf{h}_i^{prop*}}_2^2, 
\end{equation}
where $\mathbf{h}_i^{prop}$ is the predicted face pose for proposal $i$ in the proposal coordinate frame, $\mathbf{h}_i^{prop*}$ is the ground-truth face pose in the same proposal (Fig.~\ref{fig:network}, $L_{pose}$). We follow the procedure mentioned in Sec.~\ref{sec:pose_conversion} to convert ground-truth poses, $\mathbf{h}_i^{img*}$, relative to the entire image, to ground-truth pose, $\mathbf{h}_i^{prop*}$, in a proposal frame. 

\begin{figure}
    \begin{subfigure}[b]{0.475\textwidth}
        \begin{subfigure}[b]{0.495\textwidth}
            \centering
            \includegraphics[width=1\linewidth]{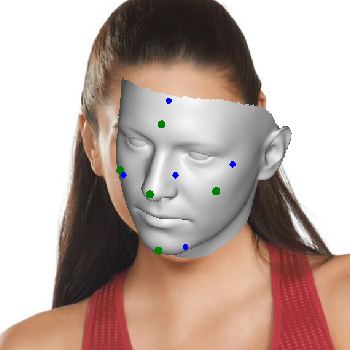}
            \caption{Wrong Pose Estimation}
        \end{subfigure}
        \hfill
            \begin{subfigure}[b]{0.495\textwidth}
            \centering
            \includegraphics[width=\linewidth]{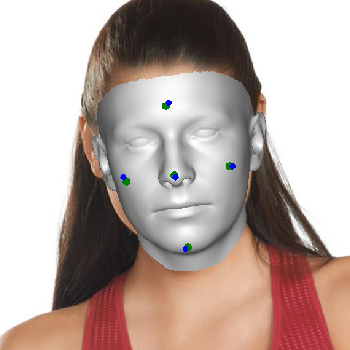}
            \caption{Correct Pose Estimation}
        \end{subfigure}
    \end{subfigure}\vspace{-2mm}
    \caption{Visualizing our calibration points. (a) When the estimated pose is wrong, points projected from a 3D face to the photo (in green) fall far from the location of these same 3D point, projected using the ground truth (in blue); (b) With a better pose estimate, calibration points projected using the estimated pose fall closer to their locations following projection using the ground-truth pose.\vspace{-2mm}}
    \label{fig:calibpoints}
\end{figure}

\minisection{(3) Calibration point loss} As an additional means of capturing the accuracy of estimated poses, we consider the 2D locations of projected 3D face shape points in the image (Fig.~\ref{fig:network}, $L_{calib}$). We compare points projected using the ground-truth pose vs. a predicted pose: An accurate pose estimate will project 3D points to the same 2D locations as the ground-truth pose (see Fig.~\ref{fig:calibpoints} for a visualization). To this end, we select a fixed set of five {\em calibration points}, $\mathbf{P}^c \in \mathbb{R}^{5 \times 3}$, on the surface of the 3D face. $\mathbf{P}^c$ is selected arbitrarily; we only require that they are not all co-planar.

Given a face pose, $\mathbf{h} \in \mathbb{R}^{6}$, either ground-truth or predicted, we can project $\mathbf{P}^c$ from 3D to 2D using Eq.~\eqref{eqn:proj}. The calibration point loss is then defined as,
\begin{equation}
    L_{calib} = \norm{\mathbf{Q}^c_i - \mathbf{Q}^{c*}_i}_1,
\end{equation}
where $\mathbf{Q}_i^c$ are the calibration points projected from 3D using predicted pose $\mathbf{h}_i^{prop}$, and $\mathbf{Q}^{c*}_i$ is the calibration points projected using the ground-truth pose $\mathbf{h}_i^{prop*}$. 

\section{Implementation details}
\subsection{Pose labeling for training and validation}\label{sec:trainingdata}
We train our method on the WIDER FACE training set~\cite{wider_face} (see also Sec.~\ref{sec:WIDER}). WIDER FACE offers manually annotated bounding box labels, but no labels for pose. The RetinaFace project~\cite{retinaface}, however, provides manually annotated, five point facial landmarks for 76k of the WIDER FACE training faces. We increase the number of training pose labels as well as provide pose annotations for the validation set, using the following weakly supervised manner.

We run the RetinaFace face bounding box and five point landmark detector on all images containing face box annotations but missing landmarks. We take RetinaFace predicted bounding boxes which have the highest IoU ratio with the ground-truth face box label, unless their IoU is smaller than $0.5$. We then use the box predicted by RetinaFace along with its five landmarks to obtain 6DoF pose labels for these faces, using standard means~\cite{posit, epnp}. Importantly, neither box or landmarks are then stored or used in our training; only the 6DoF estimates are kept. Finally, poses are converted to their global, image frames using the process described in Sec.~\ref{sec:pose_conversion}. 

This process provided us with $12,874$ images containing $138,722$ annotated training faces of which $62,827$ were assigned with weakly supervised poses. Our validation set included $3,205$ images with $34,294$ pose annotated faces, all of which were weakly supervised. During training, we ignore faces which do not have pose labels.

\minisection{Data augmentation} Similar to others~\cite{fdnet}, we process our training data, augmenting it to improve the robustness of our method. Specifically, we apply random crop, mirroring and scale transformations to the training images. Multiple scales were produced for each training image, where we define the minimum size of an image as either $640$, $672$, $704$, $736$, $768$, $800$, and the maximum size is set as $1400$. 
\subsection{Training details}\label{sec:training}
We implemented our img2pose approach in PyTorch using ResNet-$18$~\cite{resnet} as backbone. We use stochastic gradient descent (SGD) with a mini batch of two images. During training, $256$ proposals per image are sampled for the RPN loss computation and $512$ samples per image for the pose head losses. Learning rate starts at $0.001$ and is reduced by a factor of $10$ if the validation loss does not improve over three epochs. Early stop is triggered if the model does not improve for five consecutive epochs on the validation set. 
Finally, the main training took $35$ epochs. On a single NVIDIA Quadro RTX $6000$ machine, training time was roughly $4$ days.

For face pose evaluation, Euler angles are the standard metric in the benchmarks used.
Euler angles suffer from several drawbacks~\cite{trinet, quatnet}, when dealing with large yaw angles.
Specifically, when yaw angle exceeds $\pm90^{\circ}$, any small change in yaw will cause significant differences in pitch and roll (See~\cite{trinet} Sec. 4.5 for an example of this issue).
Given that the WIDER FACE dataset contains many faces whose yaw angles are larger than $\pm90^{\circ}$, 
to overcome this issue, for face pose evaluation, we fine-tuned our model on 300W-LP~\cite{3ddfa}, which only contains face poses with yaw angles in the range of $(-90, +90)$.

300W-LP is a dataset with synthesized head poses from 300W~\cite{300w} containing $122,450$ images.
Training pose rotation labels are obtained by converting the 300W-LP ground-truth Euler angles to rotation vectors, and pose translation labels are created using the ground-truth landmarks, using standard means~\cite{posit, epnp}.
During fine-tuning, $2$ proposals per image are sampled for the RPN loss and $4$ samples per image for the pose head losses.
Finally, learning rate is kept fixed at $0.001$ and the model is fine-tuned for $2$ epochs.

\begin{table*}[ht]
    \centering{
    \resizebox{0.84\linewidth}{!}{
        \begin{tabular}{l@{}crrrrrrrr}
        \toprule 
        \multicolumn{1}{c}{\textbf{Method}} & \textbf{Direct?} & \multicolumn{1}{c}{\textbf{Yaw}} & \multicolumn{1}{c}{\textbf{Pitch}} & \multicolumn{1}{c}{\textbf{Roll}} & \multicolumn{1}{c}{\textbf{MAE$_r$}} & \multicolumn{1}{c}{\textbf{X}} & \multicolumn{1}{c}{\textbf{Y}} & \multicolumn{1}{c}{\textbf{Z}} & \multicolumn{1}{c}{\textbf{MAE$_t$}} \\ \midrule
        Dlib (68 points) \cite{dlib} & {\color{red} \xmark} & 18.273 & 12.604 & 8.998 & 13.292 & 0.122 & 0.088 & 1.130 & 0.446 \\
        3DDFA \cite{3ddfa} $\dagger$ & {\color{red} \xmark} & 5.400 & 8.530 & 8.250 & 7.393 & - & - & - & - \\
        FAN (12 points) \cite{bulat2017far} $\dagger$& {\color{red} \xmark} & 6.358 & 12.277 & 8.714 & 9.116 & - & - & - & - \\
        Hopenet ($\alpha = 2$) \cite{hopenet} $\dagger$ & {\color{red} \xmark} & 6.470 & 6.560 & 5.440 & 6.160 & - & - & - & - \\
        QuatNet \cite{quatnet} $\dagger$ & {\color{red} \xmark} & 3.973 & 5.615 & 3.920 & 4.503 & - & - & - & - \\
        FSA-Caps-Fusion \cite{fsanet}  & {\color{red} \xmark} & 4.501 & 6.078 & 4.644 & 5.074 & - & - & - & - \\
        HPE \cite{hpe} $\dagger$ & {\color{red} \xmark} & 4.870 & 6.180 & 4.800 & 5.280 & - & - & - & - \\
        TriNet \cite{trinet} $\dagger$ & {\color{red} \xmark} & 4.198 & 5.767 & 4.042 & 4.669 & - & - & - & - \\
        RetinaFace R-50 (5 points) \cite{retinaface} & \textbf{\color{darkgreen} \cmark} &
        5.101 & 9.642 & 3.924 & 6.222 & 0.038 & 0.049 & 0.255 & 0.114 \\ \hline
        img2pose (ours) & \textbf{\color{darkgreen} \cmark} & \textbf{3.426} & \textbf{5.034} & \textbf{3.278} & \textbf{3.913} & \textbf{0.028} & \textbf{0.038} & \textbf{0.238} & \textbf{0.099} \\
        \bottomrule
        \end{tabular}
        }
        }\vspace{-3mm}
        \caption{Pose estimation accuracy on AFLW2000-3D~\cite{3ddfa}. $\dagger$ denotes results reported by others. Direct methods, like ours, were not tested on the ground-truth face crops, which capture scale information. Some methods do not produce or did not report translational accuracy. Finally, MAE$_r$ and MAE$_t$ are the Euler angles and translational MAE, respectively. On a $400\times400$ pixel image from AFLW2000, our method runs at 41 fps.\vspace{-5mm}}
        \label{tab:aflw_results}
\end{table*}

\begin{table}[t]
    \centering{
    \resizebox{1.0\linewidth}{!}{
        \begin{tabular}{l@{}crrrr}
        \toprule
        \textbf{Method} & \textbf{Direct?} & \textbf{Yaw} & \textbf{Pitch} & \textbf{Roll} & \textbf{MAE$_r$}
        \\ \midrule
        Dlib (68 points) \cite{dlib} $\dagger$ & {\color{red} \xmark} & 16.756 & 13.802 & 6.190 & 12.249 \\
        3DDFA \cite{3ddfa} $\dagger$ & {\color{red} \xmark} & 36.175 & 12.252 & 8.776 & 19.068 \\
        FAN (12 points) \cite{bulat2017far} $\dagger$ & {\color{red} \xmark} & 8.532 & 7.483 & 7.631 & 7.882 \\
        Hopenet ($\alpha = 1$) \cite{hopenet} $\dagger$ & {\color{red} \xmark} & 4.810 & 6.606 & 3.269 & 4.895 \\
        QuatNet \cite{quatnet} $\dagger$ & {\color{red} \xmark} & 4.010 & 5.492 & \textbf{2.936} & 4.146 \\
        FSA-NET \cite{fsanet} $\dagger$ & {\color{red} \xmark} & 4.560 & 5.210 & 3.070 & 4.280 \\
        HPE \cite{hpe} $\dagger$ & {\color{red} \xmark} & 4.570 & 5.180 & 3.120 & 4.290 \\
        TriNet \cite{trinet} $\dagger$ & {\color{red} \xmark} & \textbf{3.046} & 4.758 & 4.112 & 3.972 \\
        RetinaFace R-50 (5 pnt.) \cite{retinaface} & \textbf{\color{darkgreen} \cmark} & 4.070 & 6.424 & 2.974 & 4.490 \\\hline
        img2pose (ours) & \textbf{\color{darkgreen} \cmark} & 4.567 & \textbf{3.546} & 3.244 & \textbf{3.786} \\ 
        \bottomrule
        \end{tabular}
           } 
   }
        \vspace{-3mm}
        \caption{Comparison of the state-of-the-art methods on the BIWI dataset. Methods marked with $\dagger$ are reported by others. Direct methods, like ours, were not tested on ground truth face crops, which capture scale information. On $933\times700$ BIWI images, our method runs at 30~fps.\vspace{-6mm}}
        \label{tab:biwi_results}
\end{table}

\section{Experimental results}
\subsection{Face pose tests on AFLW2000-3D}\label{sec:AFLW2000}
AFLW2000-3D~\cite{3ddfa} contains the first $2$k faces of the AFLW dataset~\cite{aflw} along with ground-truth 3D faces and corresponding $68$ landmarks. The images in this set have a large variation of pose, illumination, and facial expression. 
To create ground-truth translation pose labels for AFLW2000-3D, we follow the process described in Sec.~\ref{sec:trainingdata}. We convert the manually annotated 68-point, ground-truth landmarks, available as part of AFLW2000-3D, to 6DoF pose labels, keeping only the translation part.
For the rotation part, we use the provided ground-truth in Euler angles (pitch, yaw, roll) format, where the predicted rotation vectors are converted to Euler angles for comparison.
We follow others~\cite{hopenet, fsanet} by removing images with head poses that are not in the range of $[-99, +99]$, discarding only $31$ out of the $2,000$ images.

We test our method and its baselines on each image, scaled to $400\times400$ pixels. Because some AFLW2000-3D images show multiple faces, we select the face that has the highest IoU between bounding boxes projected from predicted face poses and ground-truth bounding boxes, which were obtained by expanding the ground-truth landmarks. We verified the set of faces selected in this manner and it is identical to the faces marked by the ground-truth labels. 

\minisection{AFLW2000-3D face pose results} Table~\ref{tab:aflw_results} compares our pose estimation accuracy with SotA methods on AFLW2000-3D. Importantly, aside from RetinaFace~\cite{retinaface}, all other methods are applied to manually cropped face boxes and not directly to the entire photo. Ground truth boxes provide these methods with 2D face translation and, importantly, scale for either pose of landmarks. This information is {\em unavailable to our img2pose} which takes the entire photo as input. Remarkably, despite having less information than its baselines, our img2pose reports a SotA {\em MAE$_r$ of 3.913}, while running at 41 frames per second (fps) with a single Titan Xp GPU.

Other than our img2pose, the only method that processes input photos directly is RetinaFace~\cite{retinaface}. Our method outperforms it, despite the much larger, ResNet-50 backbone used by RetinaFace, its greater supervision in using not only bounding boxes and five point landmarks, but also per-subject 3D face shapes, and its more computationally demanding training. This result is even more significant, considering that this RetinaFace model {\em was used to generate some of our training labels} (Sec.~\ref{sec:trainingdata}). We believe our superior results are due to img2pose being trained to solve a simpler, 6DoF pose estimation problem, compared with the RetinaFace goal of bounding box and landmark regression.

\subsection{Face pose tests on BIWI}\label{sec:BIWI}
BIWI~\cite{biwi} contains $15,678$ frames of $20$ subjects in an indoor environment, with a wide range of face poses. 
This benchmark provides ground-truth labels for rotation (rotation matrix), but not for the translational elements required for full 6DoF.
Similar to AFLW2000-3D, we convert the ground-truth rotation matrix and prediction rotation vectors to Euler angles for comparison.
We test our method and its baselines on each image using $933\times700$ pixels resolution. Because many images in BIWI contain more than a single face, to compare our predictions, we selected the face that is closer to the center of the image with a face score $p_i > 0.9$. Here, again, we verified that our direct method detected and processed all the faces supplied with test labels. 

\minisection{BIWI face pose results} Table~\ref{tab:biwi_results} reports BIWI results following {\em protocol 1}~\cite{hopenet, fsanet} where models are trained with external data and tested in the entire BIWI dataset. Similarly to the results on AFLW2000, Sec.~\ref{sec:AFLW2000}, our pose estimation results again outperform the existing SotA, despite being applied to the entire image, without pre-cropped and scaled faces, reporting {\em MAE$_r$ of 3.786}. Finally, img2pose runtimeon the original $933\times700$ BIWI images is 30~fps.

\begin{figure*}[ht]
\begin{minipage}{0.62\textwidth}
\captionsetup{type=table}
    \begin{center}
    \resizebox{0.95\linewidth}{!}{
        \begin{tabular}{l@{}cccrrrrrr}
        \toprule
        & & & \multicolumn{3}{c}{\textbf{Validation}} & \multicolumn{3}{c}{\textbf{Test}}\\
        \multicolumn{1}{c}{\textbf{Method}} & \textbf{Backbone} & \textbf{Pose?} & \multicolumn{1}{c}{\textbf{Easy}} & \multicolumn{1}{c}{\textbf{Med.}} & \multicolumn{1}{c}{\textbf{Hard}} & \multicolumn{1}{c}{\textbf{Easy}} & \multicolumn{1}{c}{\textbf{Med.}} & \multicolumn{1}{c}{\textbf{Hard}}\\ \midrule
        \multicolumn{9}{c}{SotA methods using heavy backbones (provided for completeness)} \\ \midrule
        SRN \cite{srn} & R-50  & {\color{red} \xmark} & 0.964 & 0.953 & 0.902 & 0.959 & 0.949 & 0.897\\
        DSFD \cite{dsfd} & R-50 & {\color{red} \xmark} & 0.966 & 0.957 & 0.904 & 0.960 & 0.953 & 0.900\\
        PyramidBox++ \cite{pyramidbox} & R-50 & {\color{red} \xmark} & 0.965 & 0.959 & 0.912 & 0.956 & 0.952 & 0.909\\
        
        RetinaFace \cite{retinaface} & R-152 & \text{  }{\textbf{\color{darkgreen} \cmark}}* & 0.971 & 0.962 & 0.920 & 0.965 & 0.958 & 0.914\\
        ASFD-D6 \cite{asfd} & - & {\color{red} \xmark} & \textbf{0.972} & \textbf{0.965} & \textbf{0.925} & \textbf{0.967} & \textbf{0.962} & \textbf{0.921}\\
        \midrule
        \multicolumn{9}{c}{Fast / small backbone face detectors}\\ \hline
        Faceboxes \cite{faceboxes} & - & {\color{red} \xmark} & 0.879 & 0.857 & 0.771 & 0.881 & 0.853 & 0.774 \\
        FastFace \cite{fastface} & - & {\color{red} \xmark} & - & - & - & 0.833 & 0.796 & 0.603\\
        LFFD \cite{lffd} & - & {\color{red} \xmark} & \textbf{0.910} & 0.881 & 0.780 & 0.896  & 0.865 & 0.770\\
        RetinaFace-M \cite{retinaface}& MobileNet & \text{  }{\textbf{\color{darkgreen} \cmark}}* & 0.907 & 0.882 & 0.738 & - & - & -\\
        ASFD-D0 \cite{asfd} & - & {\color{red} \xmark} & 0.901 & 0.875 & 0.744 & - & - & -\\
        Luo \etal \cite{luoetal} & - & {\color{red} \xmark} & - & - & - & \textbf{0.902} & 0.878 & 0.528\\ \hline
        \textbf{img2pose (ours)} & R-18 & {\textbf{\color{darkgreen} \cmark}} & 0.908 & \textbf{0.899} & \textbf{0.847} & 0.900 & \textbf{0.891} & \textbf{0.839}\\
        \bottomrule
        \end{tabular}
        }\vspace{-4mm}
        \end{center}
        \caption{WIDER FACE results. '*' Requires PnP to get pose from landmarks. Our img2pose surpasses other light backbone detectors on Med. and Hard sets, despite not being trained to detect faces.}
        \label{tab:wider_results}
\end{minipage}
\hspace{3mm}
\begin{minipage}{0.35\textwidth}
\centering
\includegraphics[clip,trim=2mm 5mm 0mm 1mm,width=.81\linewidth]{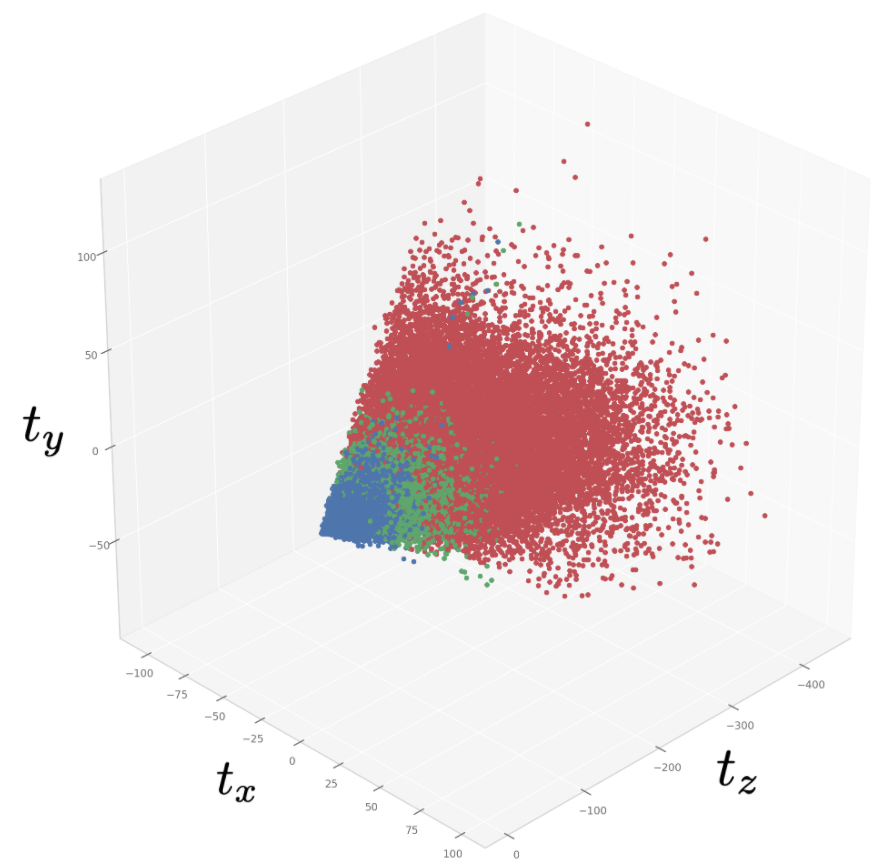}\vspace{-2mm}
    \caption{Visualizing our estimated pose translations on WIDER FACE val. images. Colors encode Easy (blue), Med. (green), and Hard (red). Easy faces seem centered close to the camera whereas Hard faces are far more distributed in the scene.}
\label{fig:analysis}
\end{minipage}
\vspace{-5mm}
\end{figure*}

\subsection{Ablation study}\label{sec:append:ablation}
We examine the effect of our loss functions, defined in Sec.~\ref{sec:posereg}, where we compare our results following training with only the face pose loss, $L_{pose}$, only the calibration points loss, $L_{calib}$, and both loss functions combined.

Table~\ref{tab:loss_ablation} provides our ablation results. The table compares the three loss variations using $MAE_r$ and $MAE_t$ on the AFLW2000-3D set and $MAE_r$ on BIWI. Evidently, combining both loss functions leads to improved accuracy in estimating head rotations, with the gap on BIWI being particularly wide, in favor of the combined loss. Curiously, translation errors on AFLW2000-3D are somewhat higher with the joint loss compared to the use of either loss function, individually. Still, these differences are small and could be attributed to stochasticity in the model training due to random initialization and random augmentations applied during training (see Sec.~\ref{sec:trainingdata} and~\ref{sec:training}).

\begin{table}[t]
    \centering
    \resizebox{0.6\linewidth}{!}{
    \begin{tabular}{lrrr}
        \toprule
        \multirow{2}{*}{\textbf{Loss}} & \multicolumn{2}{c}{\textbf{AFLW2000-3D}} & \multicolumn{1}{c}{\textbf{BIWI}} \\
         & \multicolumn{1}{c}{\textbf{MAE$_r$}} & \multicolumn{1}{c}{\textbf{MAE$_t$}} & \multicolumn{1}{c}{\textbf{MAE$_r$}}  \\
         \midrule
        $L_{pose}$ & 5.305 & \textbf{0.114} & 4.375  \\
        $L_{calib}$ & 4.746 & 0.118 & 4.023 \\
        $L_{pose}$ + $L_{calib}$ & \textbf{4.657} & 0.125 & \textbf{3.856} \\
        \bottomrule
    \end{tabular}
    }
    \caption{Comparison of the effects of different loss functions on the pose estimation results obtained on the AFLW2000-3D and BIWI benchmarks. MAE$_r$ and MAE$_t$ are the rotational and translational MAE, respectively.}
    \label{tab:loss_ablation}\vspace{-5mm}
\end{table}

\subsection{Face detection on WIDER FACE}\label{sec:WIDER}
Our method outperforms SotA methods for face pose estimation on two leading benchmarks. Because it is applied to the input images directly, it is important to verify how accurate is it in detecting faces. To this end, we evaluate our img2pose on the WIDER FACE benchmark~\cite{wider_face}. WIDER FACE offers $32,203$ images with $393,703$ faces annotated with bounding box labels. These images are partitioned into $12,880$ training, $3,993$ validation, and $16,097$ testing images, respectively. Results are reported in terms of detection mean average precision (mAP), on the WIDER FACE easy, medium, and hard subsets, for both validation and test sets.

We train our img2pose on the WIDER FACE training set and evaluate on the validation and test sets using standard protocols~\cite{retinaface, ssh_face, s3_face}, including application of flipping, and multi-scaling testing, with the shorter sides of the image scaled to  $[500, 800, 1100, 1400, 1700]$ pixels. We use the process described in Sec.~\ref{sec:method} to project points from a 3D face shape onto the image and take a bounding box containing the projected points as a detected bounding box (See also Fig.~\ref{fig:network}). Finally, box voting~\cite{gidaris2015object} is applied on the projected boxes, generated at different scales.

\begin{figure*}[!ht]
\centering{
        \includegraphics[width=1\textwidth,clip,trim=0mm 25mm 0mm 10mm]{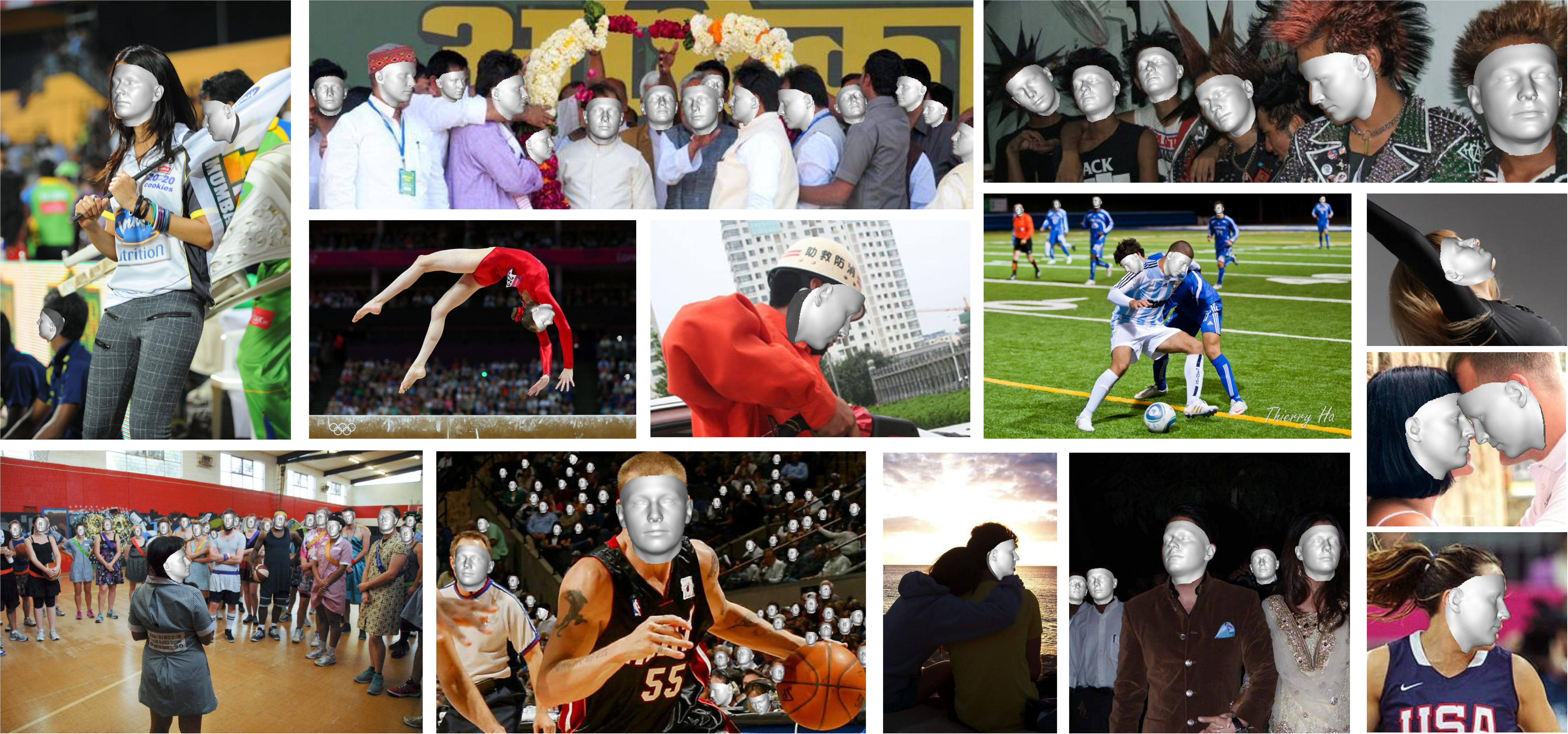}\vspace{-2mm}
    \caption{Qualitative img2pose results on WIDER FACE validation images~\cite{wider_face}. In all cases, we only estimate 6DoF face poses, directly from the photo, and without a preliminary face detection step. For more samples, please see Appendix~\ref{sec:append:qual} for more results.
    \vspace{-5mm}}\label{fig:wider_qualitative}
    }
\end{figure*}

\minisection{WIDER FACE detection results} Table~\ref{tab:wider_results} compares our results to existing methods. Importantly, the design of our img2pose is motivated by run-time. Hence, with a ResNet-18 backbone, it cannot directly compete with far heavier, SotA face detectors. Although we provide a few SotA results for completeness, we compare our results with methods that, similarly to us, use light and efficient backbones. 

Evidently, our img2pose outperforms models of comparable complexity in the validation and test,  Medium and Hard partitions. This results is remarkable, considering that our method is the only one that provides 6DoF pose and direct face alignment, and not only detects faces. Moreover, our method is trained with 20k less faces than prior work. We note that RetinaFace~\cite{retinaface} returns five face landmarks which can, with additional processing, be converted to 6DoF pose. Our img2pose, however, reports better face detection accuracy than their light model and substantially better pose estimation as evident from Sec.~\ref{sec:AFLW2000} and Sec.~\ref{sec:BIWI}.

Fig.~\ref{fig:analysis} visualizes the 3D translational components of our estimated 6DoF poses, for WIDER FACE validation images. Each $(t_x,t_y,t_z)$ point is color coded by: Easy (blue), Medium (green), and Hard (red). This figure clearly shows how faces in the easy set congregate close to the camera and in the center of the scene, whereas faces from the Medium and Hard sets vary more in their scene locations, with Hard especially scattered, which explains the challenge of that set and testifies to the correctness of our pose estimates.

Fig.~\ref{fig:wider_qualitative} provides qualitative samples of our img2pose on WIDER FACE validation images. We observe that our method can generate accurate pose estimation for faces with various pitch, yaw, roll angles, and for images under various scale, illumination, occlusion variations. These results demonstrate the effectiveness of img2pose for direct pose estimation and face detection.

\section{Conclusions}
We propose a novel approach to 6DoF face pose estimation and alignment, which does not rely on first running a face detector or localizing facial landmarks. To our knowledge, we are the first to propose such a multi-face, direct approach. We formulate a novel pose conversion algorithm to maintain consistency of poses estimated for the same face across different image crops. We show that face bounding box can be generated via the estimated 3D face pose -- achieving face detection as a byproduct of pose estimation. Extensive experiments have demonstrated the effectiveness of our img2pose for face pose estimation and face detection. 

As a class, faces offer excellent opportunities to this marriage of pose and detection: faces have well-defined appearance statistics which can be relied upon for accurate pose estimation. Faces, however, are not the only category where such an approach may be applied; the same improved accuracy may be obtained in other domains, e.g., retail~\cite{goldman2019precise}, by applying a similar direct pose estimation step as a substitute for object and key-point detection.

{\small
\bibliographystyle{ieee_fullname}
\bibliography{main}
}

\appendix

\section{Pose conversion methods}\label{sec:append:poseconvert}
\begin{figure*}[!t]
    \centering
    \begin{subfigure}{.45\textwidth}
      \centering
      \includegraphics[width=.8\linewidth]{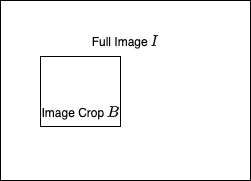}  
      \caption{An example photo}
      \label{fig:pose_photo}
    \end{subfigure}
    \begin{subfigure}{.45\textwidth}
      \centering
      \includegraphics[width=.8\linewidth]{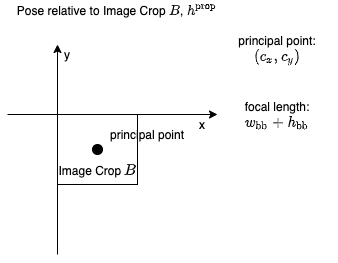}  
      \caption{Initial pose estimation $h^{prop}$}
      \label{fig:initial_pose}
    \end{subfigure}
    \newline
    \begin{subfigure}{.45\textwidth}
      \centering
      \includegraphics[width=.8\linewidth]{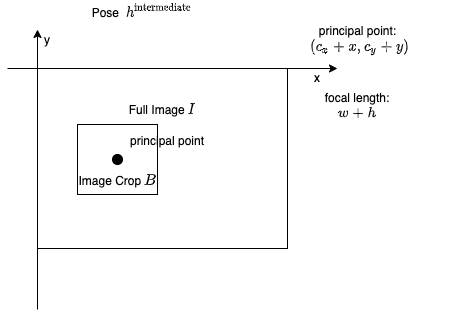}  
      \caption{Intermediate pose estimation $h^{intermediate}$ (Line 2 - 3)}
      \label{fig:intermediate_pose}
    \end{subfigure}
    \begin{subfigure}{.45\textwidth}
      \centering
      \includegraphics[width=.8\linewidth]{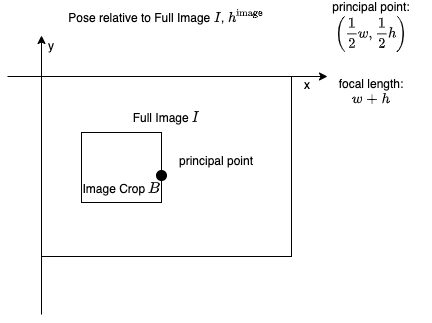}  
      \caption{Final pose estimation $h^{image}$ (Line 4 - 8)}
      \label{fig:final_pose}
    \end{subfigure}
    \caption{Illustrating the pose conversion method. See Sec.~\ref{sec:append:poseconvert} for more details.}
    \label{fig:fig}
\end{figure*}

We elaborate on our pose conversion algorithms, mentioned in Sec.~\ref{sec:pose_conversion}. Algorithm~\ref{alg:poseconversion} starts with an initial pose $\mathbf{h}^{prop}$ estimated relative to an image crop, $B$ (see, Fig.~\ref{fig:initial_pose}), and produces the final converted pose, $\mathbf{h}^{img}$, relative to the whole image, $I$ (in Fig.~\ref{fig:final_pose}).

At a high level, our pose conversion algorithm, Algorithm~\ref{alg:poseconversion}, consists of the following two steps:

The first step is a rescaling step (from Fig.~\ref{fig:initial_pose} to Fig.~\ref{fig:intermediate_pose}), where we adjust the camera to view the entire image, $I$, not just the crop, $B$.
After the first step, we obtain an intermediate pose representation, $\mathbf{h}^{intermediate}$, relative to the camera location, assumed in Fig.~\ref{fig:intermediate_pose}. 

The second step is a translation step (from Fig..~\ref{fig:intermediate_pose} to Fig.~\ref{fig:final_pose}), where we translate the principal / focal point of the camera from the center of the crop region to image center. After this step, each converted global pose, $\mathbf{h}^{img}$, from different crop, $B_i$, is estimated based on a consistent camera location, as shown in Fig.~\ref{fig:final_pose}.

Each pose, $\mathbf{h}^{prop}$, $\mathbf{h}^{intermediate}$, and $\mathbf{h}^{image}$, is associated with a specific assumed camera location and thus a specific intrinsic camera matrix, $\mathbf{K}$, $\mathbf{K}_{box}$, and $\mathbf{K}_{img}$ respectively, where we define again here. 

Here, we assume $f$ equals the image crop height, $h_{bb}$, plus width, $w_{bb}$, $c_x$ and $c_y$ are the $x,y$ coordinates of the image crop center, respectively, and $w$, $h$ are the full image width and height respectively.
\begin{equation}
    \mathbf{K} = \begin{bmatrix}
    f & 0 & c_x\\
    0 & f & c_y \\
    0 & 0 & 1
    \end{bmatrix},\label{eq:append:intrinsic}
\end{equation}
\begin{equation}
    \mathbf{K}_{box} = \begin{bmatrix}
    w+h & 0 & c_x + x\\
    0 & w+h & c_y + y\\
    0 & 0 & 1
    \end{bmatrix},\label{eq:append:cropproj}
\end{equation}
\begin{equation}
    \mathbf{K}_{img} = \begin{bmatrix}
    w+h & 0 & w/2\\
    0 & w+h & h/2 \\
    0 & 0 & 1
    \end{bmatrix}.\label{eq:append:imgproj}
\end{equation}
The input to Algorithm~\ref{alg:poseconversion} $\mathbf{h}^{prop}$ is estimated based on camera matrix, $\mathbf{K}$, whose principal point is at the center of the image crop $B$, $(c_x, c_y)$, and focal length $f$ is $w_{bb} + h_{bb}$, which is visualized in Fig.~\ref{fig:initial_pose}.
Step 1 of the algorithm, lines 2-3, first rescales the image. This {\em zoom-out} operation pushes the object further away from the camera by multiplying the translation on the $z$ axis, $t_z$, with the factor $(w+h) / (w_{bb} + h_{bb})$. This extra factor in $z$ will adjust the projected coordinates, $p$, on the image plane to reflect the relative ratio of the image crop to the whole image (since the original pose estimate $\mathbf{h}^{prop}$ is estimated assuming each image crop is of constant size).

Then we also adjust, accordingly, the camera matrices from $\mathbf{K}$ to $\mathbf{K}_{box}$. This transformation in intrinsic camera matrices will adjust the principal point, and thus the origin of the image coordinates system from the top left corner of the image crop to the top left corner of the whole image. 

Step 2 of the algorithm, lines 4-8, translates the camera, so that every pose estimate, $\mathbf{h}^{img}$, is based on the camera settings shown in Fig.~\ref{fig:final_pose} with principal point at image center and focal length $w+h$. 

The methodology here is to first adjust the camera matrix from $\mathbf{K}_{box}$ to $\mathbf{K}_{img}$, in order to compensate the translation of our desired principal points, and then solve for the associated pose, $\mathbf{h}^{img}$. Since the image coordinate system does not change in Step 2, the following equality must hold,
$$\mathbf{p} = \mathbf{K}_{box} [\mathbf{R} | \mathbf{t}] \mathbf{P},$$
$$\mathbf{p} = \mathbf{K}_{img} [\mathbf{R'} | \mathbf{t'}] \mathbf{P}.$$
In other words,
$$\mathbf{K}_{box} [\mathbf{R} | \mathbf{t}]
= \mathbf{K}_{img} [\mathbf{R'} | \mathbf{t'}].$$
So we can obtain the rotation matrices $\mathbf{R'}$ and translation vectors $\mathbf{t}'$ by the following equations,
$$\mathbf{R'} = (\mathbf{K}_{img})^{-1} \mathbf{K}_{box} \mathbf{R},$$
$$\mathbf{t'} = (\mathbf{K}_{img})^{-1} \mathbf{K}_{box} \mathbf{t}.$$
The new pose, $\mathbf{h}^{img}$, can then be extracted from $\mathbf{R'}$ and $\mathbf{t}'$ using standard approaches~\cite{rodrigues, hartley2003multiple}

The conversion from global pose, $\mathbf{h}^{img}$, to local pose, $\mathbf{h}^{prop}$, follows the exact same methodology. For completeness, we provide pseudo-code for this step in Algorithm~\ref{alg:poseconversionrev}.
\begin{algorithm}[h]
\caption{Global to local pose conversion}
\label{alg:poseconversionrev}
\begin{algorithmic}[1]
	\Procedure{pose\_convert}{$\mathbf{h}^{img}$, $\mathbf{B}$, $\mathbf{K}_{box}$, $\mathbf{K}_{img}$}
	    \State $\mathbf{V} = \mathbf{K}_{img} [t_x, t_y, t_z]^T$
	    \State $[t'_x, t'_y, t'_z]^T = (\mathbf{K}_{box})^{-1} \mathbf{V}$
	    \State $\mathbf{R} = \text{rot\_vec\_to\_rot\_mat}([r_x, r_y, r_z])$
	    \State $\mathbf{R'} = (\mathbf{K}_{box})^{-1} \mathbf{K}_{img} \mathbf{R} $
	    \State $(r'_x, r'_y, r'_z) = \text{rot\_mat\_to\_rot\_vec}(\mathbf{R'})$
	    \State $f \leftarrow w + h$
	    \State $t'_z = t'_z / f * (w_{bb} + h_{bb})$
	    \State \textbf{return} $\mathbf{h}^{prop} = (r'_x, r'_y, r'_z, t'_x, t'_y, t'_z)$  
	\EndProcedure
\end{algorithmic}
\end{algorithm}

\section{Qualitative results}\label{sec:append:qual}
\begin{figure*}[!ht]
    \centering
    \begin{subfigure}[b]{1\textwidth}
        \centering
        \includegraphics[height=0.95\textheight]{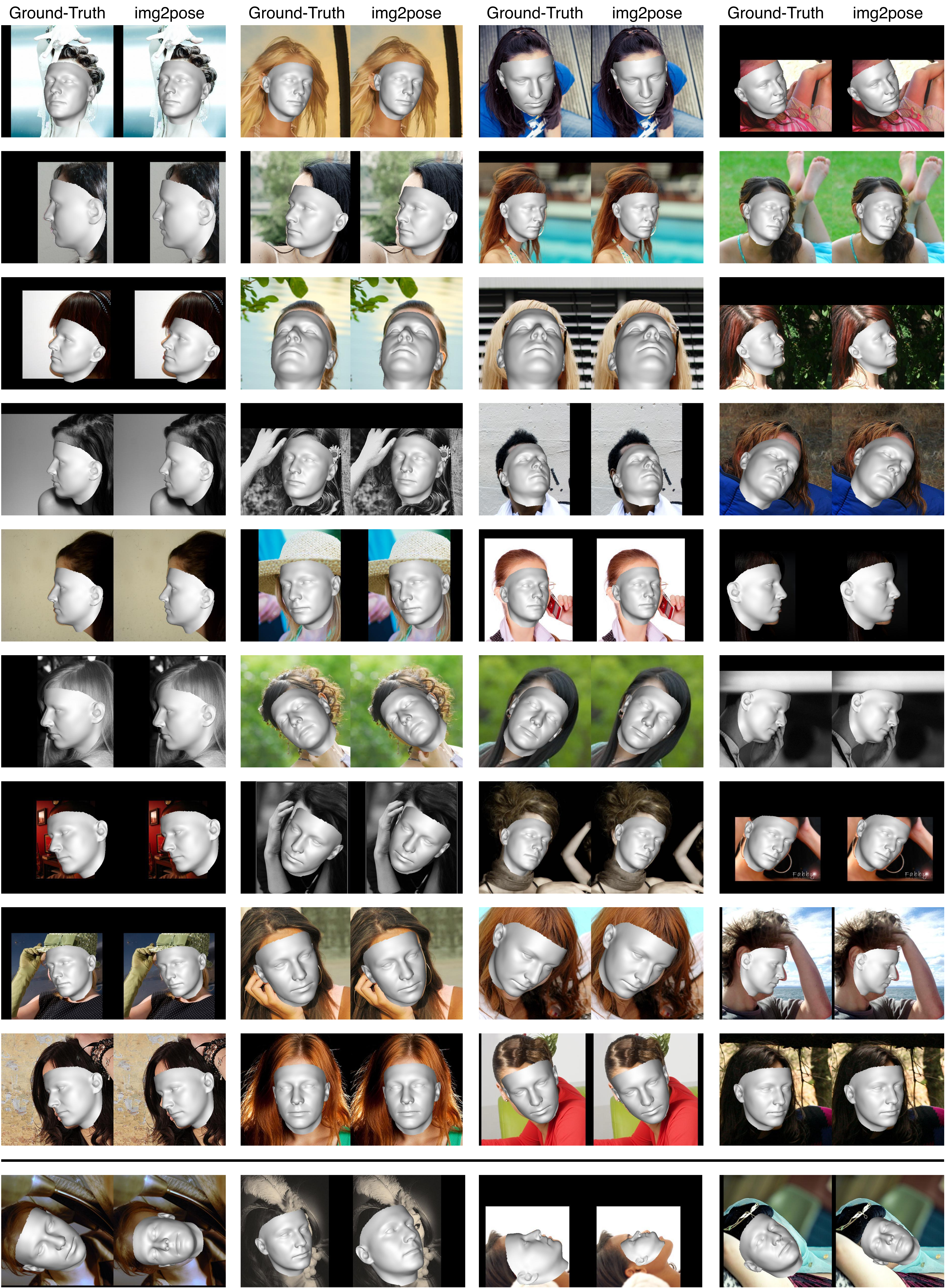}
    \end{subfigure}
    \caption{Qualitative comparing the results of our img2pose method on images from the AFLW2000-3D set to the ground truth poses. Poses visualized using a 3D face shape rendered using the pose on input photos. We provide results reflecting a wide range of face poses and viewing settings. The bottom row provides sample qualitative errors.}
    \label{fig:aflw2000_qualitative}
\end{figure*}

\begin{figure*}[!ht]
    \centering
    \begin{subfigure}[b]{1\textwidth}
        \centering
        \includegraphics[height=0.9\textheight]{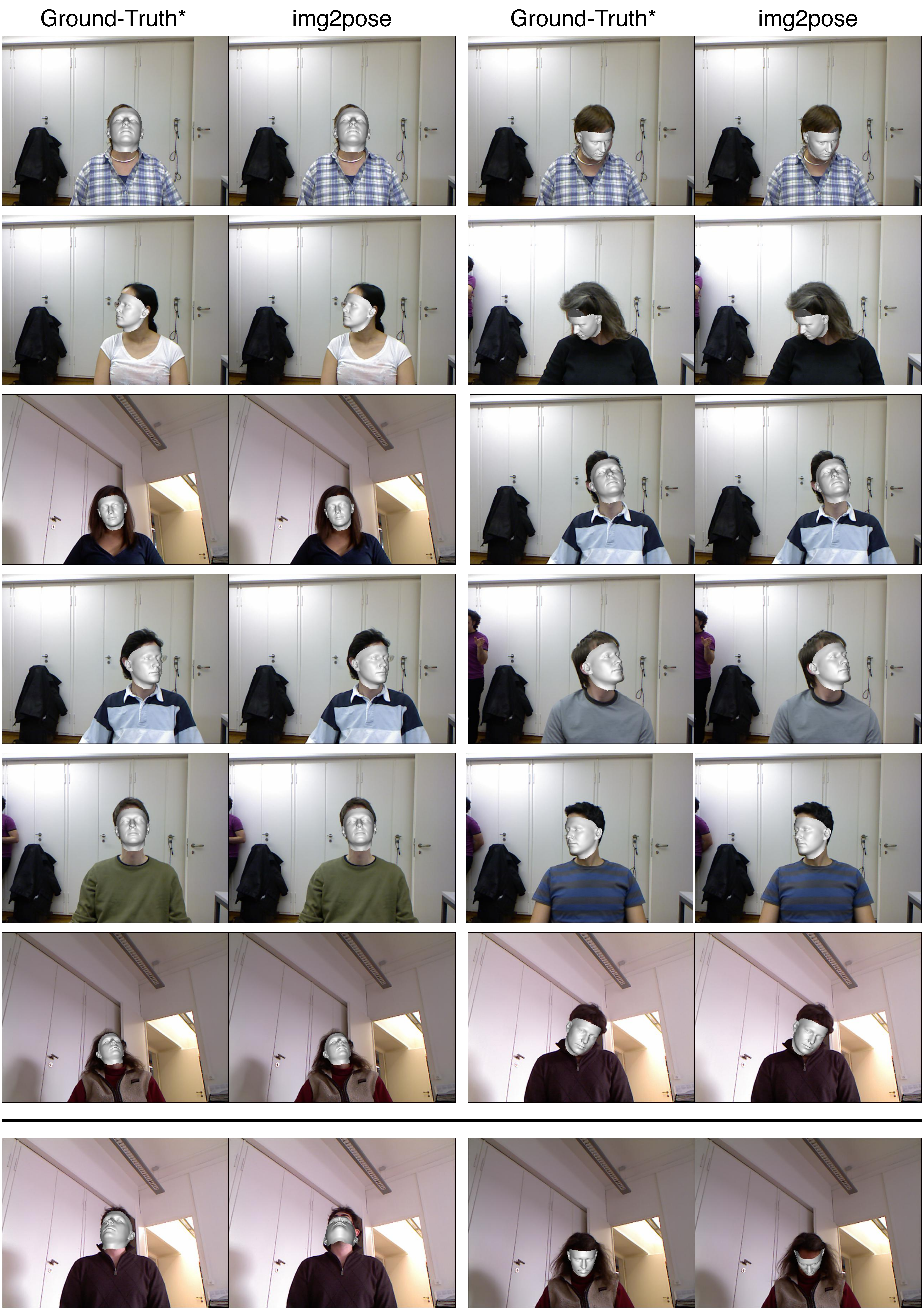}
    \end{subfigure}
    \caption{Qualitative pose estimation results on BIWI images, comparing the poses estimated by our img2pose with the ground truth. These results demonstrate how well our method correctly estimates poses for even small faces. The bottom row provides samples of the limitations of our model. Note that in all these images, the translation component of the pose, $(t_x, t_y, t_z)$, was estimated by our img2pose both for our results and the ground truth, as ground truth labels do not provide this information.}
    \label{fig:biwi_qualitative}
\end{figure*}

\begin{figure*}[!ht]
    \centering
    \begin{subfigure}[b]{1\textwidth}
        \centering
        \includegraphics[height=0.95\textheight]{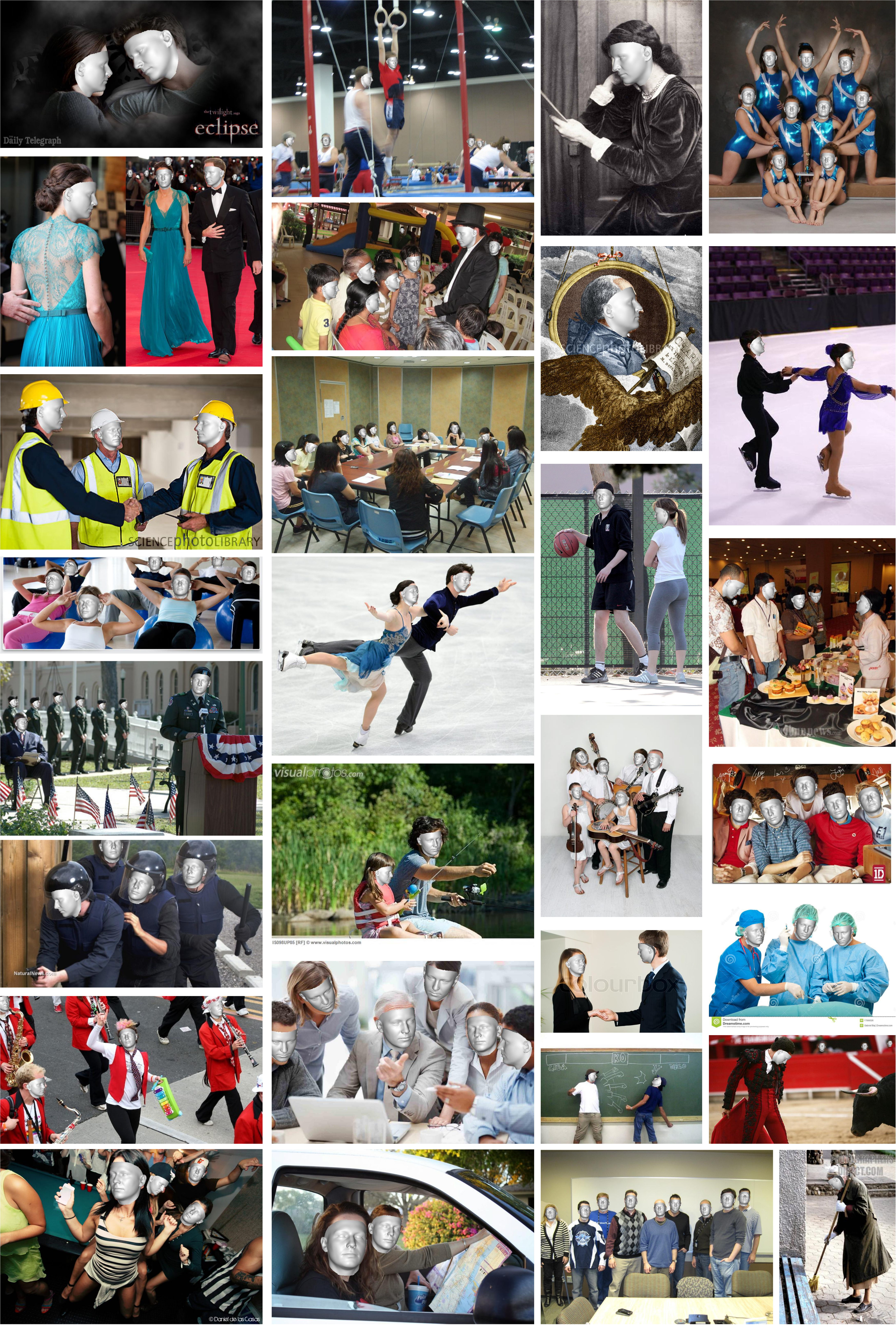}
    \end{subfigure}
    \caption{Qualitative face detection results of our img2pose method on photos from the WIDER FACE validation set. Note that despite not being directly trained to detect faces, our method captures even the smallest faces appearing in the background as well as estimates their poses (zoom in for better views).}
    \label{fig:wider_qualitative_supp}
\end{figure*}

We provide an abundance of qualitative results in Fig.~\ref{fig:aflw2000_qualitative}, ~\ref{fig:biwi_qualitative} and~\ref{fig:wider_qualitative_supp}.
Fig.~\ref{fig:aflw2000_qualitative} visually compares the pose estimated by our img2pose with the ground-truth pose labels on the AFLW2000-3D set images~\cite{3ddfa}. Our method is clearly robust to a wide range of face poses, as also evident from its state of the art (SotA) numbers reported in Table~\ref{tab:aflw_results}. The last row in Fig.~\ref{fig:aflw2000_qualitative} offers samples where our method did not accurately predict the correct pose. 

Fig.~\ref{fig:biwi_qualitative} offers qualitative results on BIWI images~\cite{biwi}, comparing our estimated poses with ground truth labels. BIWI provides ground truth angular and translational pose labels. Because we do not have information on the world (3D) coordinates used by BIWI to define their translations, we could only use their rotational ground truth values. The visual comparison should therefore only focus on the angular components of the pose. 

Our img2pose evidently predicts accurate poses, consistent with the quantitative results reported in Table~\ref{tab:biwi_results}. It is worth noting that BIWI faces are often smaller in size, relative to the entire photos, compared to the face to image sizes in AFLW2000-3D. Nevertheless, our direct method successfully predicts accurate poses. The last row of Fig.~\ref{fig:biwi_qualitative} provides sample pose errors. 

Finally, Fig.~\ref{fig:wider_qualitative_supp} provides qualitative results on the WIDER FACE validation set images~\cite{wider_face}. The images displayed show the robustness of our method across a wide range of scenarios, with varying illumination, scale, large face poses and occlusion.

\end{document}